\documentclass[runningheads]{llncs}
\usepackage{graphicx}

\usepackage{tikz}
\usepackage{comment}
\usepackage{amsmath,amssymb} 
\usepackage{color}
\usepackage{float}
\usepackage{bm}
\usepackage{caption}
\usepackage[margin=0pt]{subfig}
\usepackage[accsupp]{axessibility}  
\usepackage[pagebackref,breaklinks,colorlinks]{hyperref}

\usepackage[width=122mm,left=12mm,paperwidth=146mm,height=193mm,top=12mm,paperheight=217mm]{geometry}

\usepackage{multirow}

\newcommand{\tabincell}[2]{
\begin{tabular}{@{}#1@{}}#2\end{tabular}
}

\begin{document}
\pagestyle{headings}
\mainmatter
\def\ECCVSubNumber{961}  

\title{Dense Teacher: Dense Pseudo-Labels for Semi-supervised Object Detection} 


\titlerunning{Dense Pseudo-Labels for Semi-supervised Object Detection}
%
\author{Hongyu Zhou\inst{1,3}\thanks{Authors contributed equally to this work.} \and
	Zheng Ge\inst{1}$^{\star}$ \and
	Songtao Liu\inst{1} \and 
    Weixin Mao\inst{1,2} \and \\
    Zeming Li\inst{1} \and
    Haiyan Yu\inst{3} \and
    Jian Sun\inst{1}
}
\authorrunning{Zhou et al.}
%
\institute{$^1$MEGVII Technology \quad $^2$Waseda University \\ $^3$Harbin Institute of Technology}
\maketitle

\begin{abstract}
	To date, the most powerful semi-supervised object detectors (SS-OD) are based on pseudo-boxes, which need a sequence of post-processing with fine-tuned hyper-parameters. In this work, we propose replacing the sparse pseudo-boxes with the dense prediction as a united and straightforward form of pseudo-label. Compared to the pseudo-boxes, our Dense Pseudo-Label (DPL) does not involve any post-processing method, thus retaining richer information. We also introduce a region selection technique to highlight the key information while suppressing the noise carried by dense labels. We name our proposed SS-OD algorithm that leverages the DPL as Dense Teacher. On COCO and VOC, Dense Teacher shows superior performance under various settings compared with the pseudo-box-based methods. Code is available at \url{https://github.com/Megvii-BaseDetection/DenseTeacher}
	\end{abstract}

\section{Introduction}

Current high-performance object detection neural networks rely on a large amount of labeled data to ensure their generalization capability. However, labeling samples takes a high cost of human effort. Thus the industry and academia pay extensive attention to the use of relatively easy-to-obtain unlabeled data. An effective way to use these data is Semi-Supervised Learning (SSL), where at the training time, only part of the data is labeled while the rest are unlabeled. On image classification tasks, the dominated method of mining information from unlabeled data is ``Consistency-based Pseudo-Labeling''~\cite{simclr,fixmatch,mixmatch,remixmatch}. Pseudo-Labeling~\cite{pseudolabel} is a technique that utilizes trained models to generate labels for unlabeled data. Meanwhile, the Consistency-based regularization~\cite{consistencyreg}, from another perspective, forces a model to have similar output when giving a normal and a perturbed input with different data augmentations and perturbations like Dropout~\cite{dropout}. 

\begin{figure}[t]
	\centering
	\includegraphics[width=120mm]{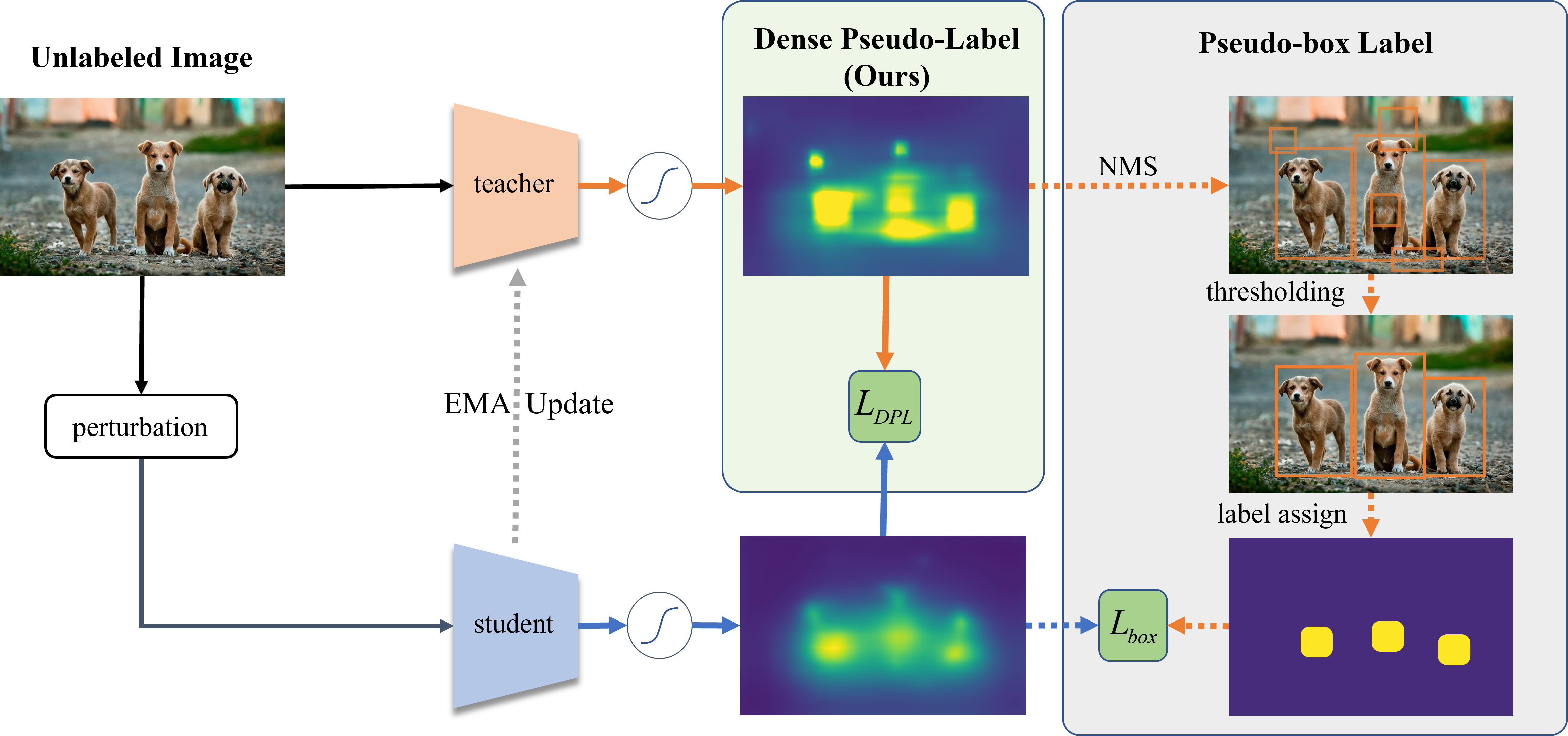}
	\caption{The overview of our purposed pipeline for unlabeled data compared with traditional pseudo-box based pipeline. For each iteration, Dense Pseudo-Label (DPL) are generated by the teacher model on unlabeled images. The student model then calculates unsupervised loss on perturbed images and corresponding DPLs. By removing post-processing steps, DPL contains rich information from the teacher model. Note that the vanilla learning approach uses only labeled data (not plotted in the figure) and the total loss is the sum of both supervised and unsupervised loss.  }
	\label{fig:1}
\end{figure}

This pipeline has been successfully transferred to Semi-Supervised Object Detection (SS-OD)~\cite{stac,ubteacher,softteacher}. Specifically, the predicted boxes from a pre-trained ``teacher'' detector are used as the annotations of unlabeled images to train the ``student'' detector, where the same images are applied with different augmentations for the teacher and student model. This instinctive method has proven to be effective in SS-OD and has achieved state-of-the-art scores on benchmarks such as COCO~\cite{coco} and Pascal VOC~\cite{voc}. However, it is not reasonable to replicate all the empirics directly from the classification task. While the generated pseudo-label is a single and united class label for an image in classification, the object detectors predict a set of pseudo-boxes as the annotation of an image. As shown in Fig.~\ref{fig:1}, making direct supervision on the unlabeled image with these pseudo-boxes requires several additional steps, including Non-Maximum-Suppression (NMS), Thresholding, and Label Assignment. Such a lengthy label-generating procedure introduces many hyper-parameters, such as NMS threshold $\sigma_{nms}$ and score threshold $\sigma_t$, substantially affecting the SS-OD performance.\footnote{See also in Sec.~\ref{sec:background} for a related discussion.} This motivates us to explore a more simple and effective form of pseudo-labels for SS-OD. 


In this work, we propose a new SS-OD pipeline named Dense Teacher with a \textit{united} form of pseudo-label --- Dense Pseudo-Label (DPL), which enables more efficient knowledge transfer between the teacher and student models. DPL is an integral label. Different from the existing box-like labels in a human-readable form, it is the original output from the network without ordinary post-processing. 
Our Dense Teacher, following existing Pseudo-Labeling paradigms, works in the following way: For each iteration, labeled and unlabeled images are randomly sampled to form a data batch. The ``teacher'' model, which is an Exponential Moving Average of the ``student'' model, generates DPL for unlabeled data. And the student model is trained on both ground truth on labeled images and DPL on unlabeled images. Since the DPL does not require any post-processing, the pipeline of Dense Teacher is extremely simple. The overall pipeline of Dense Teacher can be seen in Fig.~\ref{fig:1}.

Although DPL provides more information than pseudo-box labels, it contains high-level noise (\emph{e.g.}, low-scoring predictions) as well. We show in Sec.~\ref{sec:analy} that learning to make those low-scoring predictions can distract the student model, resulting in poor detection performance. Therefore, we propose a region division method to suppress noise and highlight key regions where the student model should concentrate on. According to our experiments, the region division strategy can effectively utilize the rich information contained in hard negative regions to enhance training. As a result, our proposed Dense Teacher, together with the region division strategy, shows state-of-the-art performance on MS-COCO and Pascal VOC.

Our main contributions in this paper are:
\begin{itemize}
	\item We conduct a thorough analysis on the drawbacks of pseudo-boxes in the SS-OD task.
	\item We propose a united form of pseudo-label named DPL to better fit the semi-supervised setting and the Dense Teacher framework to apply the DPL on one-stage detector.
	\item The proposed Dense Teacher achieves state-of-the-art performance on MS-COCO and Pascal VOC benchmarks under various settings. Gain analysis  and ablation study are provided to verify the effectiveness of each part in Dense Teacher.
\end{itemize}

\section{Related Works}
\subsection{Semi-Supervised Learning}
Semi-Supervised Learning (SSL) means that a portion of the data is labeled at training time while the other is not. 
Currently, there are two main approaches to achieve this goal: pseudo-labeling and consistency regularization. Pseudo-label-based methods~\cite{pseudolabel} first train a network on labeled data, then use the trained network as a teacher to make inferences on unlabeled data. This prediction result is then assigned to a specific class according to a threshold based on the predicted confidence and used as labeled data to train another student network. In~\cite{mean_teacher}, the teacher model is replaced by the Exponential Moving Average (EMA) of the student to conduct online pseudo-labeling. 
Consistency regularization based methods~\cite{consistencyreg} construct a regularization loss to force predictions under a set of perturbations ${\lbrace T_i \rbrace}$ to be same. Perturbations can be implemented using augmentation~\cite{simclr,perturbreg,noisystudent}, dropout~\cite{temporalensemble}), or adversarial training~\cite{adversarial}. This approach does not require annotation and can be used in combination with other methods; therefore, it is widely adopted in many SSL frameworks~\cite{pseudoensemble,fixmatch,mixmatch}.

\subsection{Object Detection}
Object detectors can be divided into Anchor-based and Anchor-free paradigms. Anchor-based detectors predict the offsets and scales of target boxes from predefined anchor boxes. Although this approach has succeeded on many tasks, one needs to redefine new anchor boxes when applying such models to new data. 
In contrast to anchor-based detectors, predefined anchor boxes are not required for anchor-free detectors. These detectors directly predict the box size and location on the feature map. Take FCOS model as an example; this detector predicts the classification score, distances to four boundaries, and a quality score on each pixel of Feature Pyramid Network (FPN)~\cite{fpn}. A variety of subsequent improvements, such as the adaptive label assigning while training~\cite{atss}, boundary distribution modeling~\cite{gflv1} were proposed to improve its performance. Considering the wide application, streamlined architecture, and excellent performance of FCOS, we will conduct our experiments under this framework.

\subsection{Semi-Supervised Object Detection}
The label type is the main difference between Semi-Supervised Object Detection (SS-OD) and SSL. Previous studies have transferred a great deal of experience from SSL works to the SS-OD domain. 
CSD~\cite{csd} use a flipped image $I^{\prime}$ to introduce a consistency loss between $F(I)$ and $F(I^{\prime})$, this regularization can be applied to unlabeled image. 
STAC~\cite{stac} train a teacher detector on labeled images and generate pseudo-labels on unlabeled data using this static teacher. These pseudo-labels will then be selected and used for training like labeled data. 
Unbiased Teacher~\cite{ubteacher} use thresholding to filter pseudo-labels, Focal Loss~\cite{focalloss} is also applied to address the pseudo-labeling bias issue. 
Adaptive Class-Rebalancing~\cite{acr} artificially adds foreground targets to images to achieve inter-class balance. \emph{Li, et al.}~\cite{rethinkingp} propose dynamic thresholding and loss re-weighting for each category. 
Soft Teacher~\cite{softteacher} proposed a score-weighted classification loss and box jittering approach to select and utilize regression loss of pseudo-boxes. 
While these methods successfully transferred paradigms from SSL to SS-OD, they ignored the unique characteristics of SS-OD. These pseudo-box-based strategies treat pseudo-boxes, or selected pseudo-boxes, as ordinary target boxes, and thus they invariably follow the detector's label assign strategy.

\section{Dense Teacher}
In Sec.~\ref{sec:plframe}, we first introduce the existing Pseudo-Labeling SS-OD framework. Then we analyze the disadvantages of utilizing pseudo-boxes in Sec.~\ref{sec:background}. In the remaining part, we propose Dense Pseudo-Label to overcome the issues mentioned above and introduce our overall pipeline in detail, including the label generation, loss function, and the learning region selection strategy. Since our primary motivation is to show the superiority of dense pseudo-labels compared to pseudo-box labels, we naturally choose to verify our idea on dense detectors (\emph{i.e.}, one-stage detectors). 


\subsection{Pseudo-Labeling Framework}
\label{sec:plframe}

Our Dense Teacher follows the existing pseudo-labeling framework~\cite{ubteacher,softteacher} as shown in Fig.~\ref{fig:1}. Within each iteration:

\begin{enumerate}
    \item Labeled and unlabeled images are randomly sampled to form a data batch.
    \item The teacher model, an exponential moving average (EMA) of the student, takes the augmented unlabeled images to generate pseudo-labels.
    \item The student model then takes the labeled data for vanilla training and calculates supervised loss $\mathcal{L}_s$, while the unlabeled data together with pseudo-labels are used to produce unsupervised loss $\mathcal{L}_u$.
    \item Two losses are weighted and learned to update parameters of the student model. The student model updates the teacher model in an EMA manner.
\end{enumerate}

\noindent Finally, the overall loss function is defined as:
\begin{gather}
	\mathcal{L} = \mathcal{L}_s + w_u \mathcal{L}_u,
\end{gather}

\noindent where $w_u$ is the unsupervised loss weight. Traditionally, the unsupervised loss $\mathcal{L}_u$ is calculated with pseudo-boxes. However, in the following section, we point out that using processed boxes as pseudo-labels can be inefficient and sub-optimal.

\subsection{Disadvantages of Pseudo-box Labels}\label{sec:background}

In this part, we study the behavior of pseudo-box-based SS-OD algorithms on COCO~\cite{coco}, as well as CrowdHuman\footnote{CrowdHuman is a benchmark for detecting humans in a crowded situation, performance is measured by Log-average Miss Rate (mMR). The lower the better.}~\cite{crowdhuman} since the impact of the NMS threshold can be more clearly demonstrated in the crowd situation. We adopt Unbiased Teacher~\cite{ubteacher} as a representative algorithm to FCOS for these experiments.

\begin{figure}[!t]
	\centering
	\begin{tabular}{@{}c@{}c@{}}
	    \subfloat[Detection Performance v.s. $\sigma_t$]{\includegraphics[width=55mm]{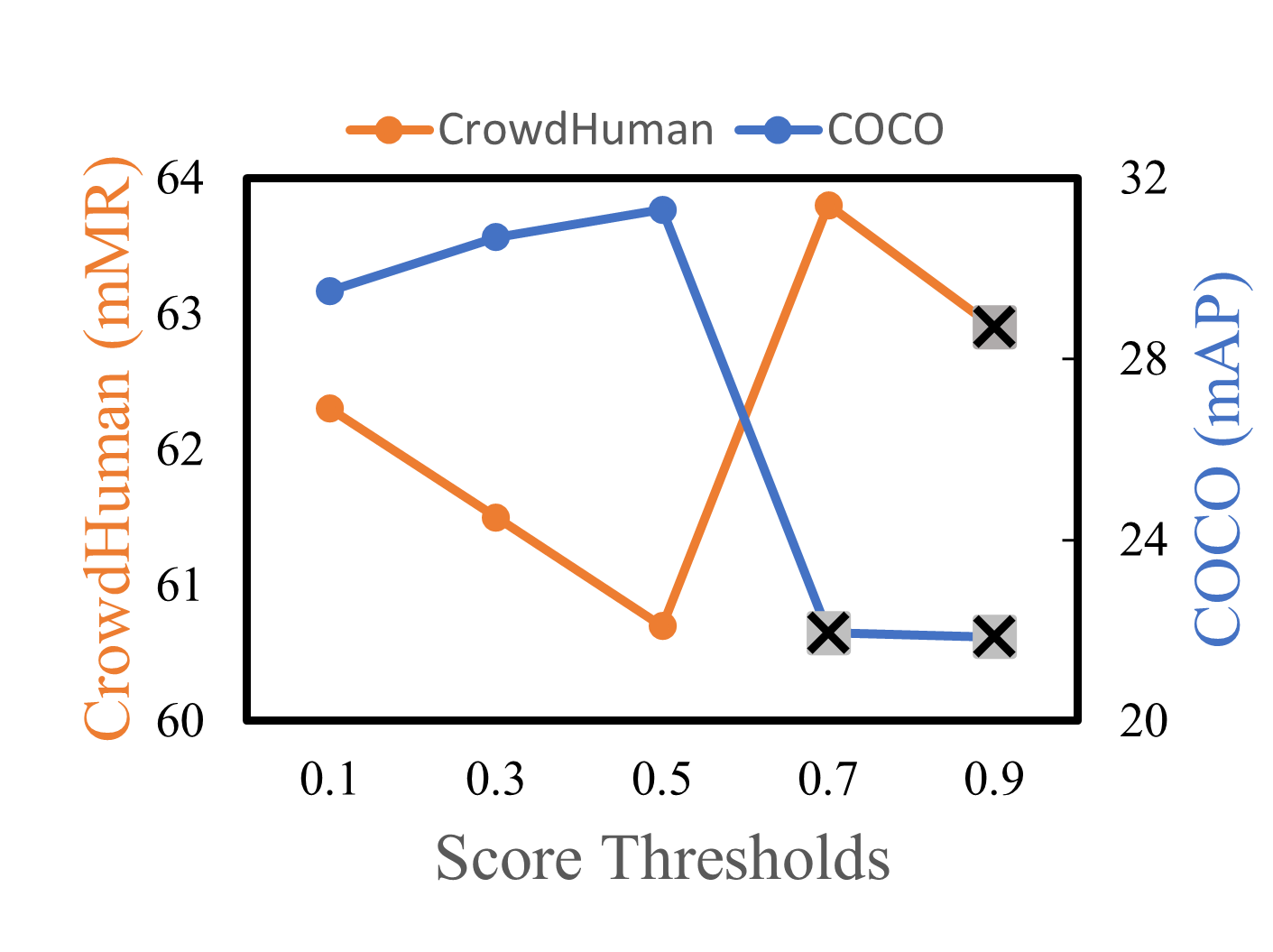}} & \subfloat[Detection Performance v.s. $\sigma_{nms}$]{\includegraphics[width=55mm]{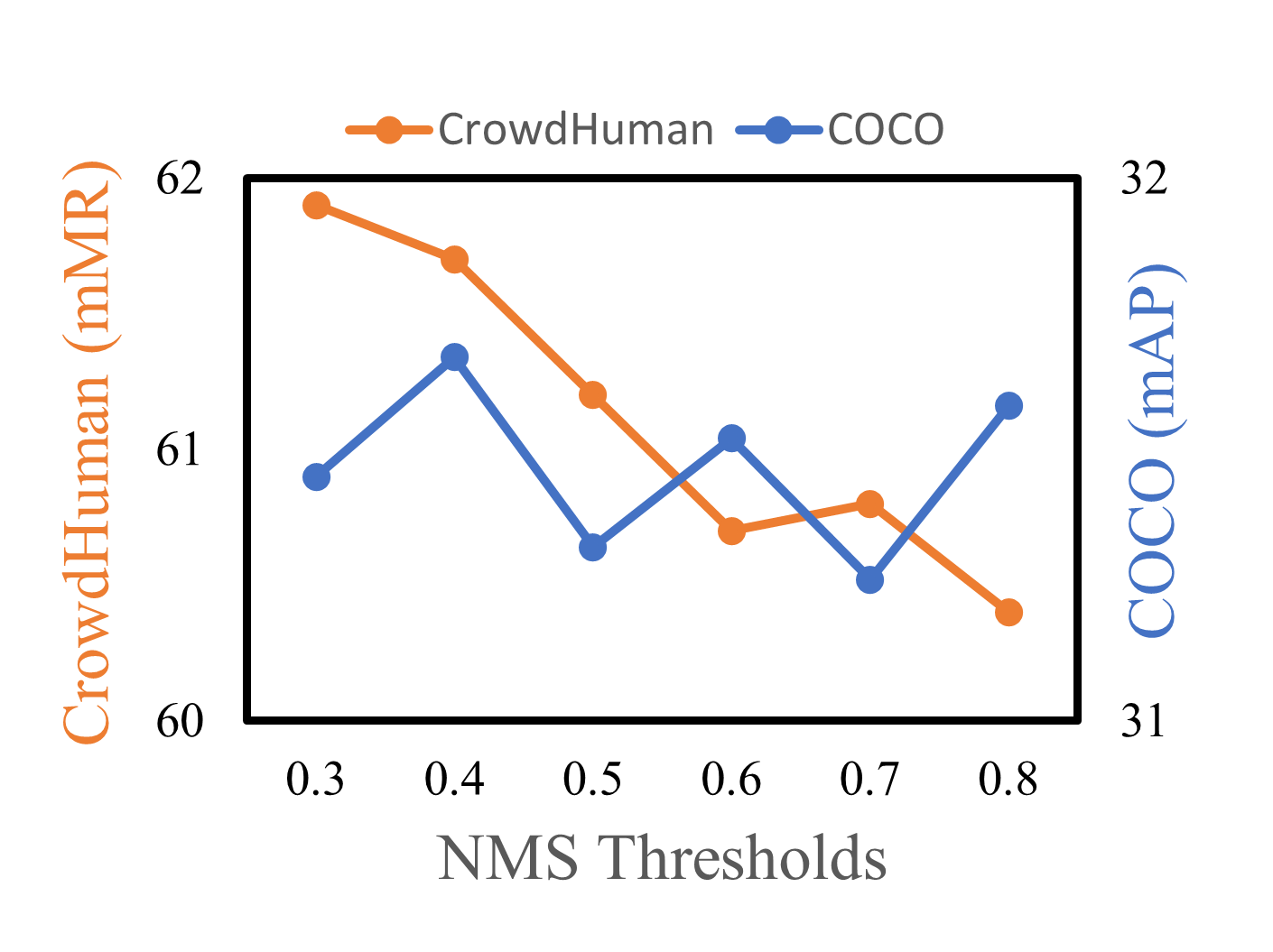}} \\
	    \subfloat[FP/FN on COCO]{\includegraphics[width=50mm]{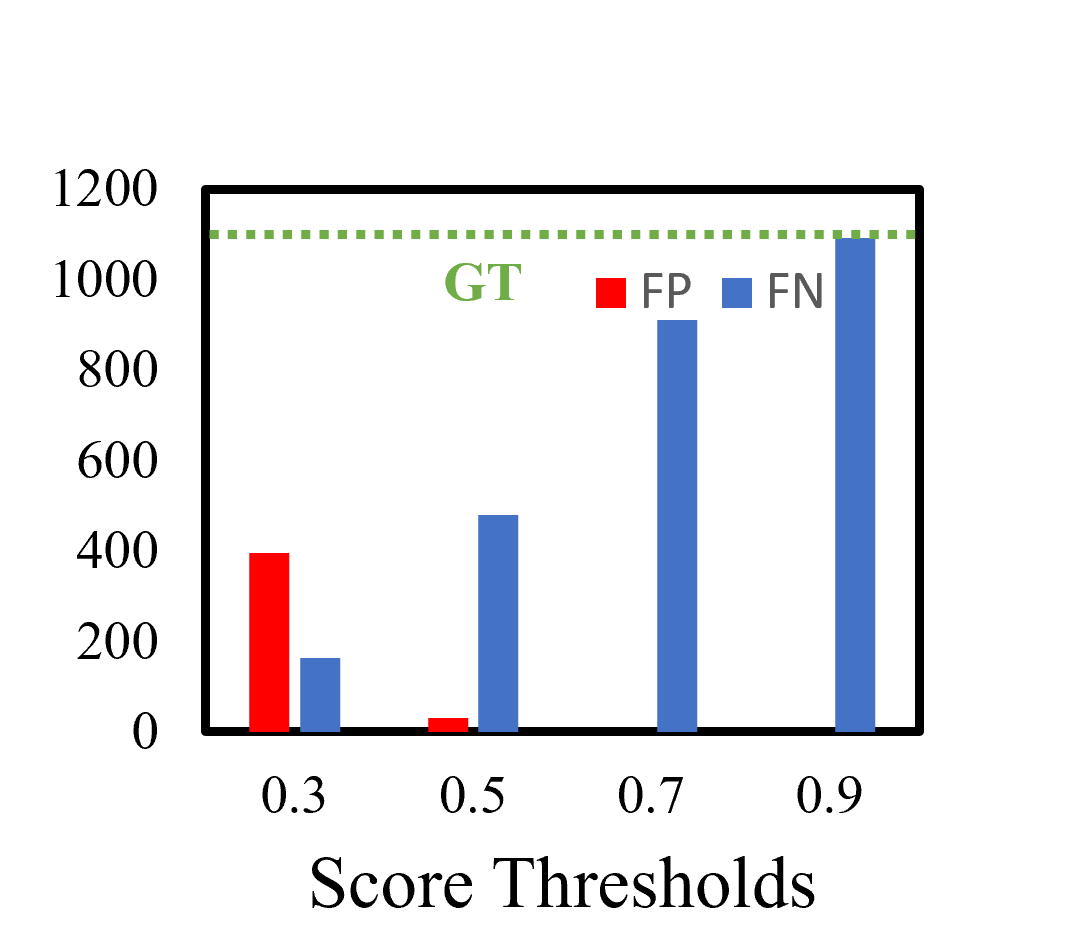}} & 
	    \subfloat[FP/FN on CrowdHuman]{\includegraphics[width=50mm]{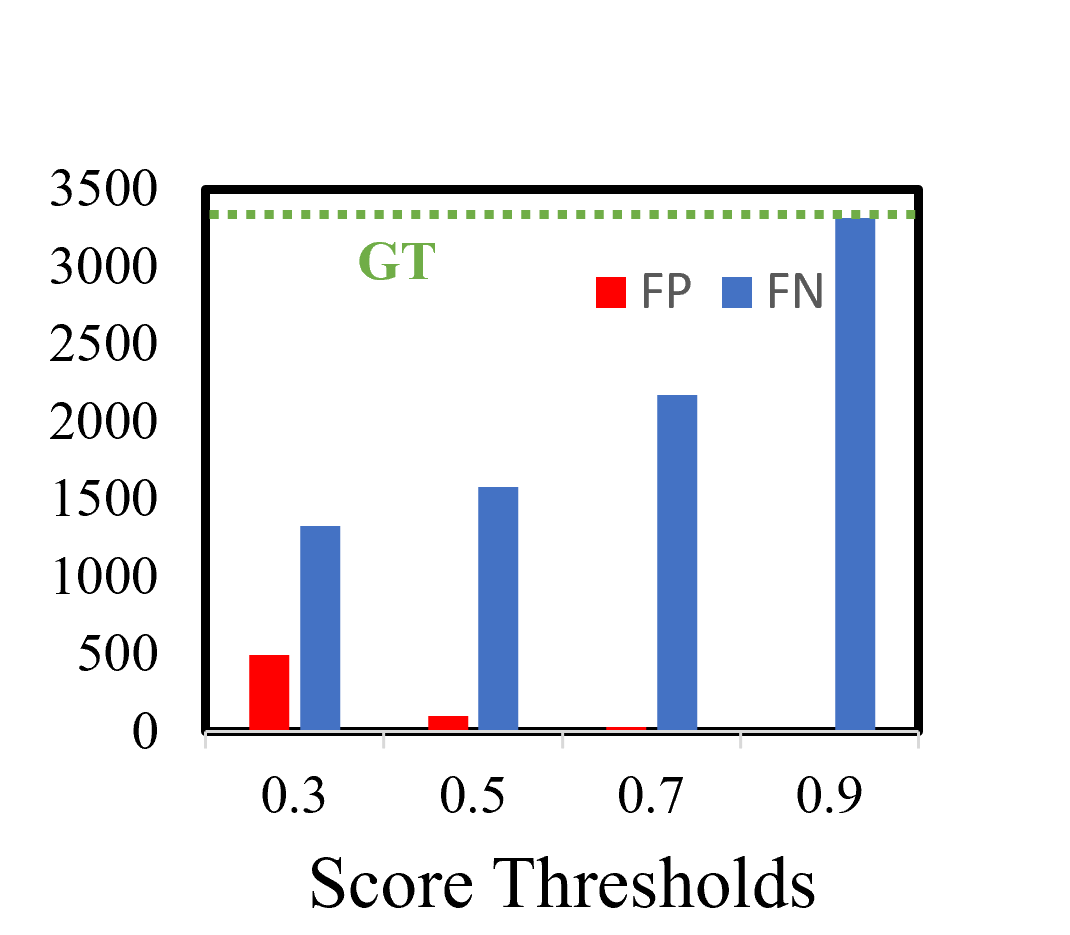}}
	\end{tabular}
	\caption{Analysis of Pseudo-box based approaches. (a) and (b): Performances under different $\sigma_{t}$ and $\sigma_{\rm NMS}$. Note that the gray $\boxtimes$ represents the training fails to converge. (c) and (d): False Positive and False Negative boxes on 128 images under different threshold on COCO and CrowdHuman, the green line denotes the ground truth box number}
	\label{fig:2}
\end{figure}

\subsubsection{Dilemma in Thresholding}
In SS-OD algorithms~\cite{ubteacher,softteacher}, the output of the teacher model is expected to play the role of ground-truth labels for unsupervised images. To this end, Thresholding is a key operation to screen out low-scoring boxes so that the quality of pseudo-box labels can be improved. However, our preliminary experiments show that the threshold $\sigma_t$ introduced by this operation may substantially affect the entire training process. In Fig.~\ref{fig:2}(a), we present the training results of Unbiased Teacher under different $\sigma_t$. 
It shows that the detection performance fluctuates significantly on both datasets as the $\sigma_t$ varies. Moreover, when $\sigma_t$ is set to a high value (\emph{e.g.}, 0.7 and 0.9), the training process even fails to converge. Such a phenomenon is possibly caused by a large number of false negatives in the teacher's prediction, as shown in Fig.~\ref{fig:2}(c) and (d). When this is the case, Thresholding will eliminate many high-quality predictions and mislead the learning process of the student model. Conversely, when set $\sigma_t$ to a low value such as 0.3, the performance shows apparent degradation due to the increasing number of false positives (see in Fig.~\ref{fig:2}(c) and (d) as well). As a result, one can not find a perfect threshold to ensure the quality of generated pseudo-boxes.

\subsubsection{Dilemma in Non-Maximum Suppression (NMS)}
NMS is adopted on the detector's original outputs for most object detection algorithms to remove redundant predictions. It is also indispensable to the teacher model in existing SS-OD frameworks, without which the resulting pseudo-labels will be a mess. NMS introduces a threshold $\sigma_{nms}$ to control the degree of suppression. According to our experiments, we find that $\sigma_{nms}$ also has a non-negligible effect on the SS-OD algorithms. Fig.~\ref{fig:2}(b) shows the relationship between $\sigma_{nms}$ and performance of Unbiased Teacher. From this figure, we can tell that 1). different $\sigma_{nms}$ may lead to fluctuations in the detection performance (especially on CrowdHuman). 2). the optimal $\sigma_{nms}$ values for different datasets are different (\emph{i.e.}, 0.7 on COCO and 0.8 on CrowdHuman), which will bring in extra workload for developers to tune the optimal $\sigma_{nms}$ on their custom datasets. Moreover, previous works~\cite{ps-rcnn,r2nms} show that in a crowd scene like in the CrowdHuman dataset, there does not exist a perfect $\sigma_{nms}$ that can keep all true positive predictions while suppressing all false positives. As a result, with NMS adopted, the unreliability of pseudo-box labels is further exacerbated.

\subsubsection{Inconsistent Label Assignment}
\begin{figure}[!t]
	\centering
	\begin{tabular}{@{}ccc@{}}
	    \subfloat[Raw Image]{\includegraphics[width=40mm]{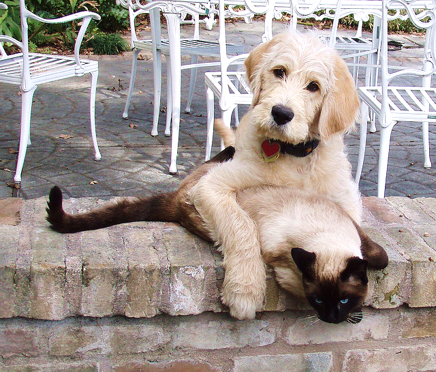}} & 
	    \subfloat[Ground Truth Positives]{\includegraphics[width=40mm]{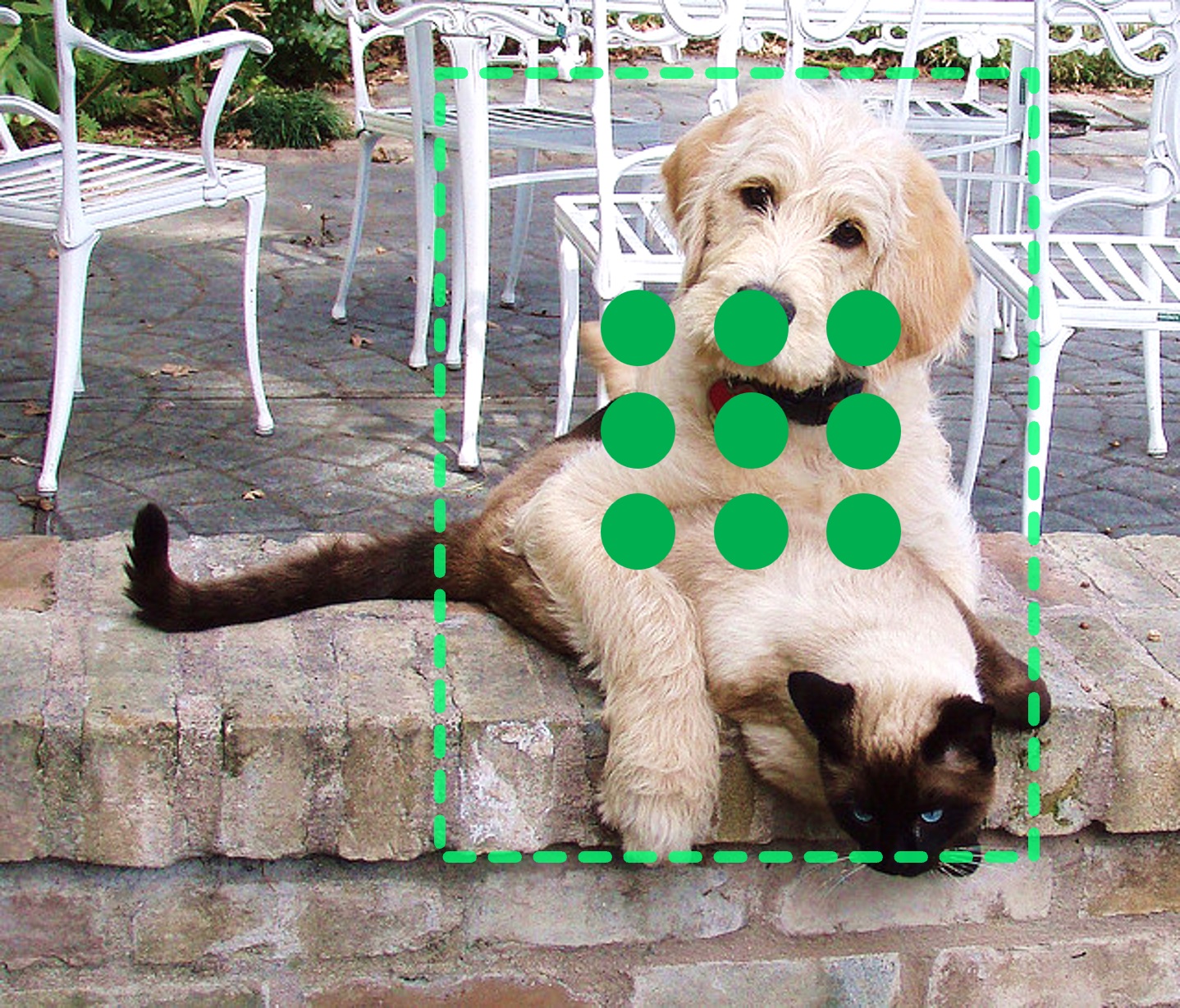}} & 
	    \subfloat[Pseudo-Box Positives]{\includegraphics[width=40mm]{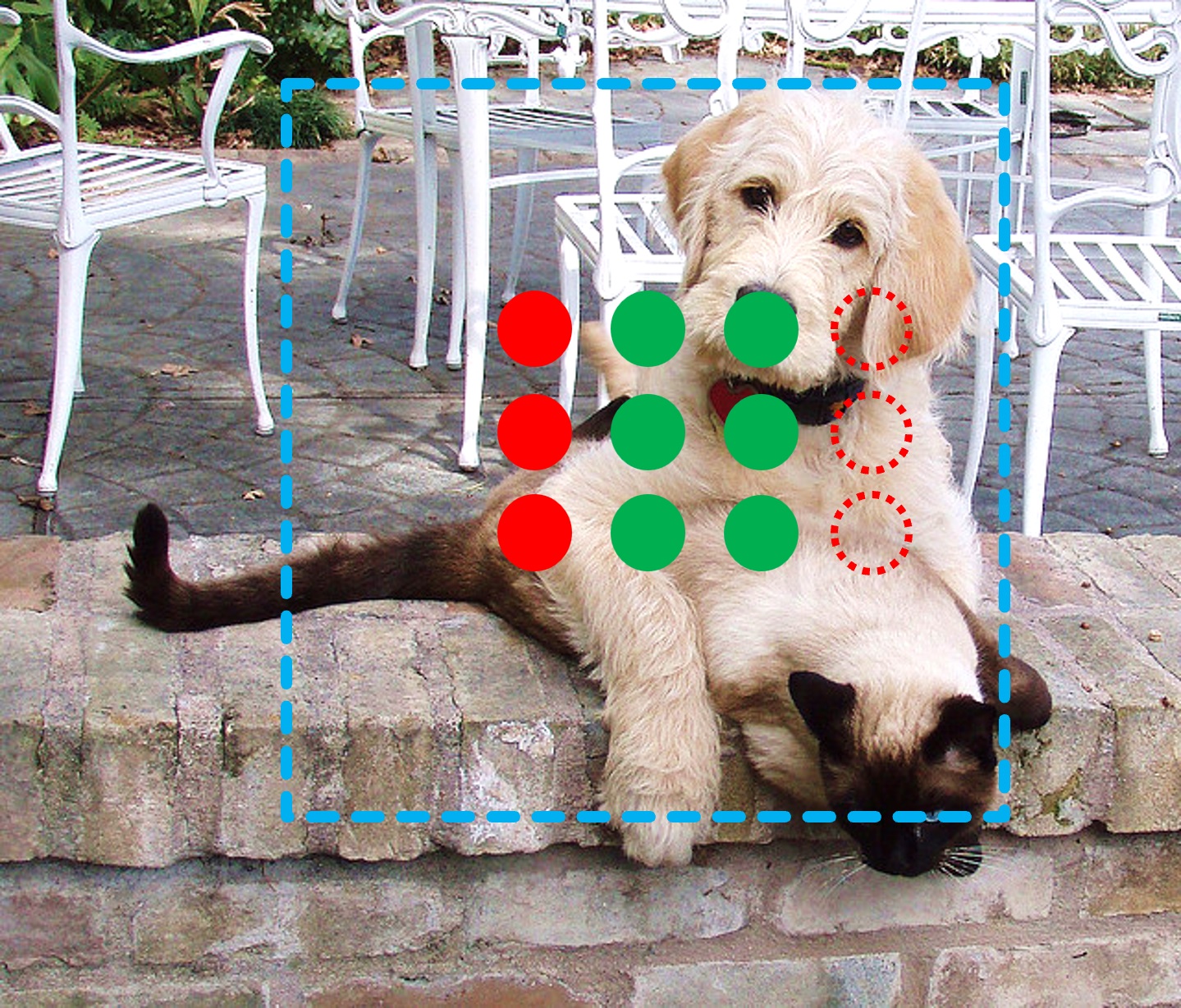}}
	\end{tabular}
	\subfloat{\includegraphics[height=6mm]{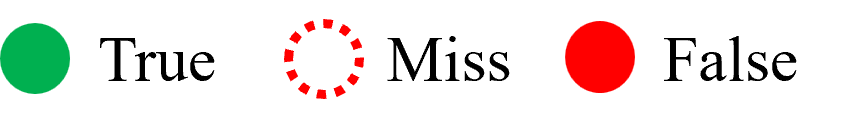}}
	\caption{Comparisons between (b) foreground pixels assigned by ground truth boxes and (c) foreground pixels assigned by pseudo-boxes}
	\label{fig:3}
\end{figure}

As shown in Fig.~\ref{fig:1}, existing pseudo-label-based algorithms convert the sparse pseudo-boxes into a dense form by label assignment to form the final supervision. An anchor box (or point) will be assigned as either positive or negative during label assignment based on a particular pre-defined rule. Although this process is natural in the standard object detection task, we believe it is harmful to SS-OD tasks. The reason is quite simple: the pseudo-boxes may suffer from the inaccurate localization problem, making the label assigning results inconsistent with the potential ground-truth labels. In Fig.~\ref{fig:3}, we can find that although the predicted box matches the actual box under IoU threshold 0.5, a severe inconsistent assigning result appears due to the inaccurate pseudo-box. This inconsistency with the ground truth is likely to degrade the performance.

Due to the above three issues, we challenge the convention of using pseudo-box as the middle-ware of unsupervised learning and propose a new form of pseudo-label that is dense and free of post-processing.

\setlength{\tabcolsep}{1.4pt}
\setlength{\tabcolsep}{4pt}

\subsection{Dense Pseudo-Label}
\label{sec:roi}

To address the problems mentioned above, we propose Dense Pseudo-Label (DPL) that encompasses richer and undistorted supervising signals. Specifically, we adopt the post-sigmoid logits predicted by the trained model as our desired dense pseudo-label, as shown in the green box in Fig.~\ref{fig:1}. After bypassing those lengthy post-processing methods, one can naturally discover that our proposed DPL reserves more detailed information from the teacher than its pseudo-box counterpart.

Since DPL represents information in continuous values (value between 0 and 1) and the standard Focal Loss~\cite{focalloss} can only deal with discrete binary values (0 or 1), we adopt Quality Focal Loss~\cite{gflv1} to conduct learning between dense pseudo-labels and the student's predicting results. Let us denote $\vec{y_i}\triangleq \vec{p^t_{i}}$ as DPL (\emph{i.e.}, teacher's prediction) and denote $\vec{p^s_{i}}$ as student's prediction for $i$-th anchor\footnote{``Anchor'' stands for ``anchor point'' in anchor-free detectors and ``anchor box'' in anchor-based detectors.}, we hope the prediction and the target to be similar for the same anchor. Therefore, we can write the classification loss on the $i$-th anchor for an unlabeled image as:
\begin{gather}
	\mathcal{L}_i^{cls}  = -  |\vec{y_i} - \vec{p^s_{i}}|^{\gamma}
	* \left[ \vec{y_i} log(\vec{p^s_{i}}) + (1-\vec{y_i}) log (1-\vec{p^s_{i}}) \right]
\end{gather}
where $\gamma$ is the suppression factor. 

While DPL contains rich information, it also keeps many low-scoring predictions due to the absence of the thresholding operation. Since those low-scoring predictions usually involve the background regions, intuitively, the knowledge encompassed in them shall be less informative. In Sec.~\ref{sec4_4}, we experimentally prove that learning to mimic the teacher's response in those regions will hurt the SS-OD algorithm's performance. Therefore, we propose to divide the whole input image into a learning region and a suppressing region(e.t., negative region in positive-negative division) based on the teacher's Feature Richness Score (FRS~\cite{frs}). With the help of this richness score, we select the pixels with top $k$\% scores as the learning region and the other regions will be suppressed to 0. As as result, our DPL is extended to:

\begin{gather}
	S_i = \max_{c \in [1, C]}(p^t_{i,c})\\
	\vec{y_i} = \left\{
	\begin{aligned}
		\vec{p^t_i},\ \  & \rm{if\ }{S_i}\rm{\ in\ top\ k\%}, \\
		\vec{0},\ \  & \rm{otherwise}.
	\end{aligned}
	\right.
\end{gather}
where $p^t_{i,c}$ denotes the score prediction of $c$-th class for $i$-th sample from the teacher, $C$ denotes the total number of classes.


Besides, this design has other advantages: 
\begin{enumerate}
    \item By modifying the learning region, we can easily achieve Hard Negative Mining by selecting extra samples (see Fig.~\ref{fig:4}). In Sec.~\ref{sec:HN} we will analyze the gain from this part in detail.
    \item Since the learning region is selected, unsupervised learning for regression branch can be easily achieved. We apply IoU Loss on this branch and analyze its gain in Sec.~\ref{sec:analy}
\end{enumerate}


\section{Experiments}


\subsection{Datasets and Experiment Settings}
\subsubsection{Datasets.}
We present our experimental results on MS-COCO~\cite{coco} and Pascal VOC~\cite{voc} benchmarks. For MS-COCO, both labeled and unlabeled training datasets will be used. The \texttt{train2017} set contains 118k images with target bounding boxes and the \texttt{unlabeled2017} contains 123k unlabeled images. Validation is performed on the subset \texttt{val2017}. For Pascal VOC, training set uses \texttt{VOC07 train} and \texttt{VOC12 train} and validation set uses \texttt{VOC07 test}. The following three experimental settings are mainly studied: \\
\begin{itemize}
	\item {\bf COCO-Standard: } 1\%, 2\%, 5\% and 10\% of the \texttt{train2017} set are sampled as labeled data, respectively. The rest of images are viewed as unlabeled data while training. 
	For fairness of comparison, we follow the same dataset division as in~~\cite{ubteacher} which contains 5 different data folds. Mean score of all 5 folds are taken as the final performance.
	\item {\bf COCO-Full: } \texttt{train2017} is used as labeled data while \texttt{unlabeled2017} is used as unlabeled data.
	\item {\bf VOC Mixture:} \texttt{VOC07 train} is used as labeled data, while \texttt{VOC12 train} and \texttt{COCO20cls}\footnote{\texttt{COCO20cls} is the sampled COCO \texttt{train2017} set, only 20 classes same as in VOC are included.} are taken as unlabeled data.
\end{itemize}
\subsubsection{Implementation Details.}
Without loss of generality, we take FCOS~\cite{fcos} as the representative anchor-free detector for experiments. ResNet-50~\cite{resnet} pre-trained on ImageNet~\cite{imagenet} is used as the backbone. We use batch-size 16 for both labeled and unlabeled images. The base learning rate and $\gamma$ in QFL are set to 0.01 and 2 in all of our experiments. Loss weight $w_u$ on unlabeled data is set to 4 on COCO-Standard and 2 on the other settings. Following previous works~\cite{ubteacher,stac}, we adopt ``burnin'' strategy to initialize the teacher model, same data augmentations as in~\cite{ubteacher} are applied.

\subsection{Main Results}
\label{sec:analy}
\begin{table}[t]
	\begin{center}
		\caption{Performance under different model configurations on COCO-Standard 10\%. * denotes our re-implemented result on FCOS. ``Our Division'' means the learning/suppression region division based on FRS score}
		\label{table:1}
		\begin{tabular}{cccccccccc}
			\hline
			Method & Learning Region & Cls          & Reg          & AP                               & AP50       & AP75      \\
			\hline
			Supervised & -
			       & -     & -     & 26.44                            & 42.69      & 28.11         \\
			\hline
			$\rm{Unbiased~Teacher}^*$~\cite{ubteacher} & Predicted Positive
			       & $\checkmark$ & $\times$     & 31.52      & 48.80      & 33.57          \\
			Dense Teacher & All
			       & $\checkmark$ & $\times$     & 32.00 & 50.29 & 34.17  \\
			Dense Teacher& Our Division
			       & $\checkmark$ & $\times$     & \textbf{33.34} & $ 52.14  $ & $ 35.53  $  \\
			\hline
			$\rm{Unbiased~Teacher}^*$~\cite{ubteacher}& Predicted Positive
			       & $\checkmark$ & $\checkmark$ & 33.13        & $49.96$    & $35.36$   \\
			Dense Teacher& Our Division
			       & $\checkmark$ & $\checkmark$ & \textbf{35.11} &  53.35  &  37.79    \\
			\hline
		\end{tabular}
	\end{center}
\end{table}


In this section, we progressively improve the Dense Teacher and analyze the performance gain from each part in detail. We adopt Unbiased Teacher as our baseline. Results on the COCO-Standard 10\% setting are shown in Table.~\ref{table:1}. We first replace Unbiased Teacher's pseudo-boxes with our proposed Dense Pseudo-Labels without the region division strategy. It shows that this improves the mAP from 31.52\% to 32.0\%. Then, we apply our region division strategy on DPL and the mAP is further improved by 1.34\%. Finally, we extend the unsupervised learning scheme to the regression branch as done by~\cite{softteacher}, and our final mAP comes to 35.11\%. To the best of our knowledge, this is the new state-of-the-art under the COCO-Standard 10\% setting. According to these results, we can attribute the advantages of Dense Teacher over existing methods to two major improvements:

\begin{figure}[bt]
	\centering
	\begin{tabular}{@{}c@{}c@{}cc}
	    \subfloat[Ground Truth]{\includegraphics[height=36mm]{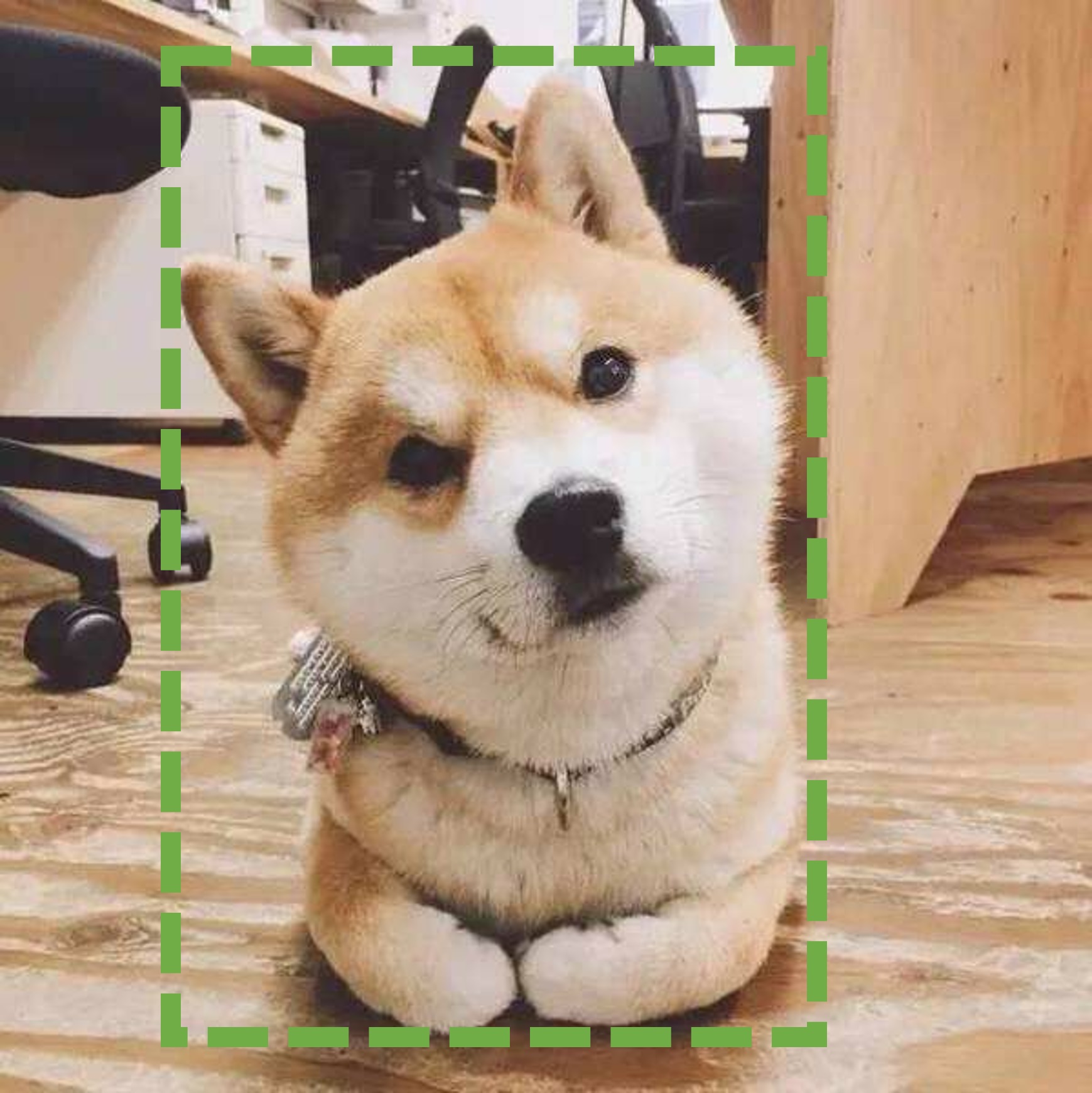}} & 
	    \subfloat[Pseudo-Box Label]{\includegraphics[height=36mm]{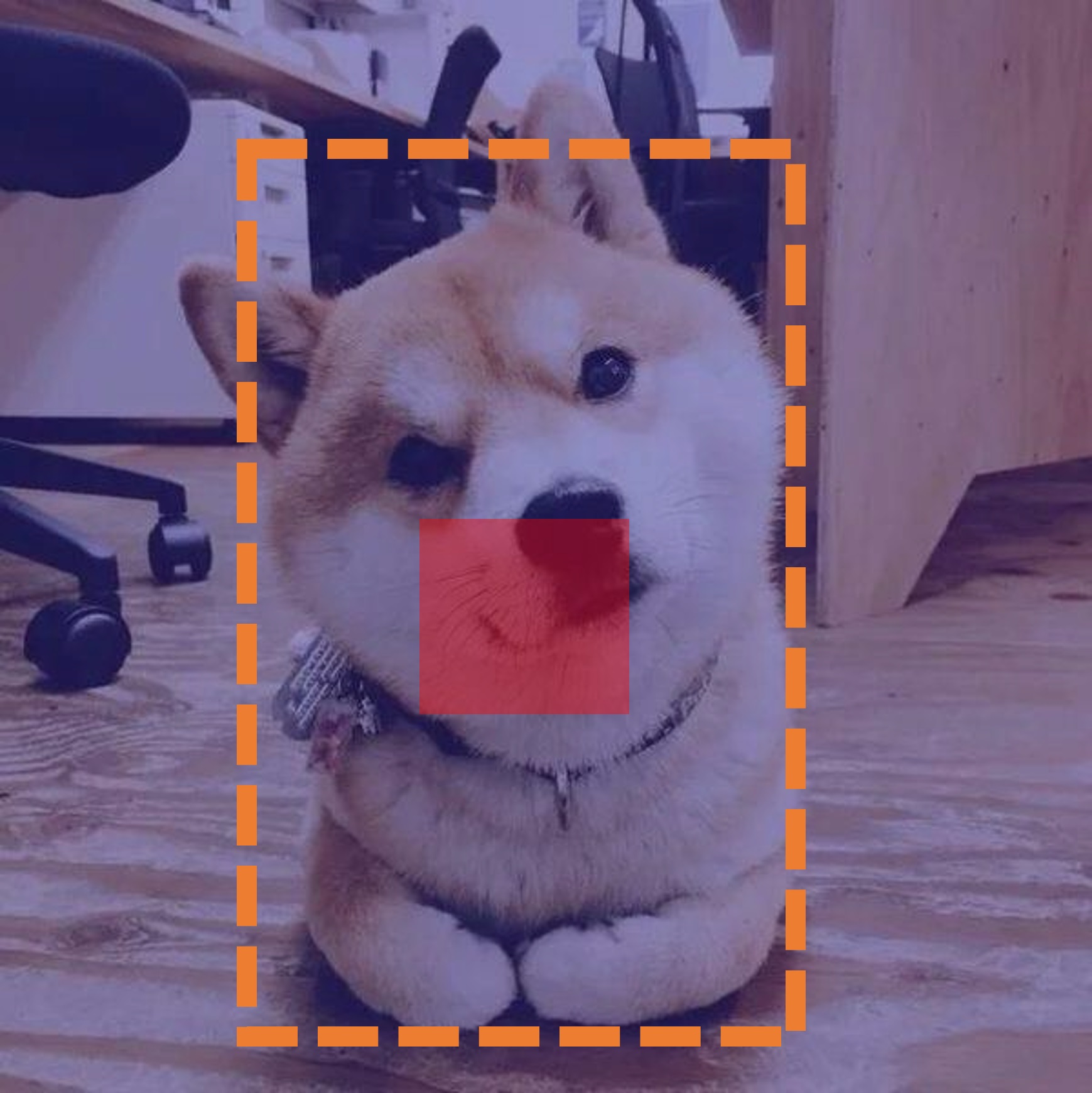}} &
	    \subfloat[Dense Pseudo-Label]{\includegraphics[height=36mm]{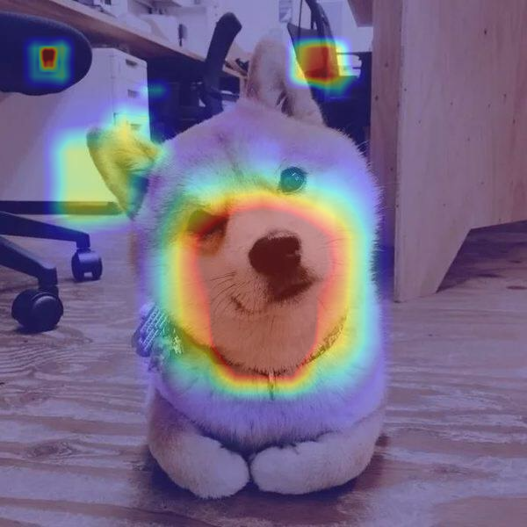}} &
	    \subfloat{\includegraphics[width=9mm]{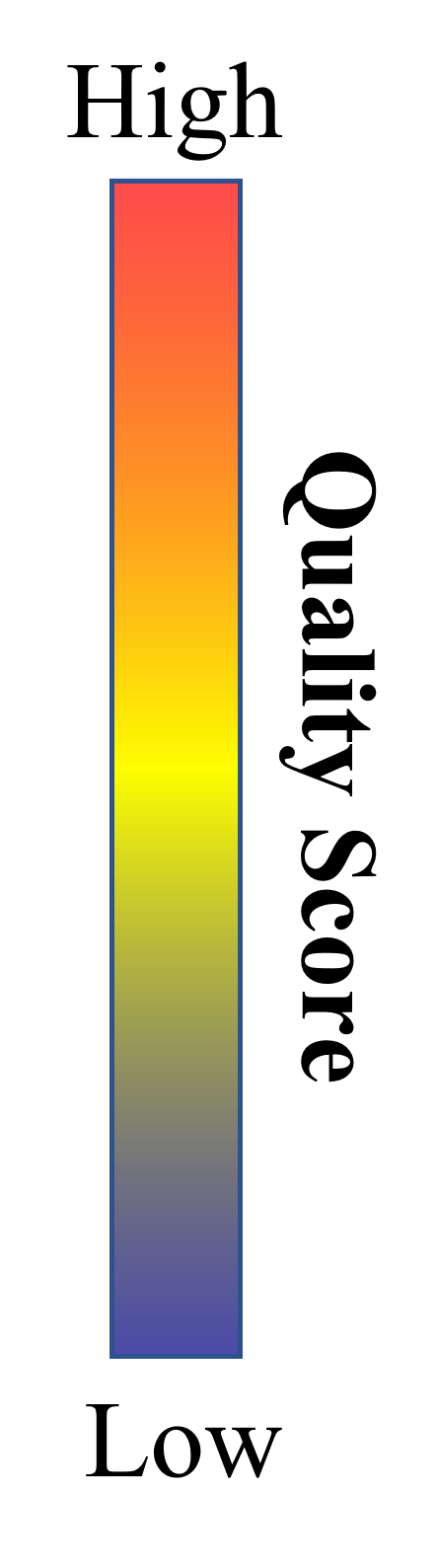}}
	\end{tabular}
	\caption{Illustration of (c) our Dense Pseudo-Label compared with (b) pseudo-box label and its assigning result. Blue areas denote the assigned negative samples. In Dense Pseudo-Label, red means high quality scores, which denotes positive samples in pseudo-box label. It can be seen that Dense Pseudo-Label is able to leverage more hard negative regions compared to the Pseudo-Box based method}
	\label{fig:4}
\end{figure}

1). The new form of pseudo-label resolves the deterioration problem of the pseudo-box label as mentioned in Sec.~\ref{sec:background}. It is worth mentioning that by getting rid of the lengthy post-processing procedure, our Dense Teacher forms a much simpler SS-OD pipeline but still with better performance. However, the resulting improvement (from 31.52\% to 32.00\%) remains marginal without the advanced learning region division strategy.

2). Our region division strategy can efficiently utilize hard negative regions to enhance training.
Specifically, we conduct label assignment on FCOS using ground truth annotation of COCO, finding that there are only about 0.4\% of positive samples in the COCO \texttt{train2017} set. By specifying $k=1$, we take a fair amount of hard negative samples for unsupervised training. 
In Fig.~\ref{fig:4}, we can see that hard negatives samples distribute on meaningful background objects (chair cushion, cabinet, and other parts of the dog) in the image. These responses from the teacher are valuable in improving the student's performance. 


\subsection{Comparison with State-of-the-arts}


\subsubsection{COCO-Standard.}

\setlength{\tabcolsep}{5pt}
\begin{table}[bth]
	\begin{center}
		\caption{Experimental results on COCO-Standard. * means our re-implemented results on FCOS, $\circledcirc$ means large scale jittering is adopted when training}
		\label{table:2}
		\scalebox{0.9}{
		\begin{tabular}{ccccc}
			\hline
			 & \multicolumn{4}{c}{COCO-Standard}\\
			\cline{2-5}
			 & 1\%                   & 2\%                & 5\%                & 10\%             \\
			\hline
			Supervised
			 & $ 11.24\pm0.18  $ & $ 15.04\pm0.31  $ & $ 20.82\pm0.13 $ & $ 26.44\pm0.11$\\
			\hline
			CSD~\cite{csd}
			 & $10.51\pm0.06$        & $13.93\pm 0.12$    & $18.63\pm 0.07$    & $22.46\pm 0.08$  \\
			STAC~\cite{stac}
			 & $ 13.97\pm 0.35 $     & $ 18.25\pm 0.25  $ & $ 24.38\pm 0.12  $ & $ 28.64\pm 0.21$ \\
			Instant Teaching~\cite{instantteaching}
			 & $18.05 \pm 0.15$ & $ 22.45\pm 0.30  $ & $ 26.75\pm 0.05  $ & $ 30.40\pm 0.05$ \\
			ISMT~\cite{meanteachers}
			 & $18.88 \pm 0.74$ & $ 22.43\pm 0.56  $ & $ 26.37\pm 0.24  $ & $ 30.52\pm 0.52$ \\
			Unbiased Teacher~\cite{ubteacher}
			 & \bm{$20.75 \pm 0.12$} & $ 24.30\pm 0.07  $ & $ 28.27\pm 0.11  $ & $ 31.50\pm 0.10$ \\
			Humble Teacher~\cite{humbleteacher}
			 & $16.96 \pm 0.38$ & $ 21.72\pm 0.24  $ & $ 27.70\pm 0.15  $ & $ 31.61\pm 0.28$ \\
			\emph{Li, et al.}~\cite{rethinkingp}
			 & $19.02 \pm 0.25$ & $ 23.34\pm 0.18  $ & $ 28.40\pm 0.15  $ & $ 32.23\pm 0.14$ \\
			\hline
			$\rm{Unbiased~Teacher}^*$~\cite{ubteacher}
			 & $18.31\pm0.44 $  & $22.39\pm 0.26$ &$ 27.73 \pm0.13 $           & $31.52\pm0.15$  \\
			Ours
			 & $19.64\pm0.34$& \bm{$25.39\pm0.13$}& \bm{$30.83\pm0.21$}        & \bm{$35.11\pm0.12$}      \\
			\hline
			$\rm{Soft~Teacher^\circledcirc}$~\cite{softteacher}
			 & $20.46\pm 0.39 $      & $ -  $             & $30.74\pm 0.08 $   & $34.04\pm 0.14$  \\
			$\rm{Ours^\circledcirc}$
			 & \bm{$22.38\pm0.31$}  & \bm{$27.20\pm0.20$}& \bm{$ 33.01\pm0.14 $} & \bm{$37.13\pm 0.12$}     \\
			\hline
		\end{tabular}
		}
	\end{center}
\end{table}
\setlength{\tabcolsep}{4pt}

We compare Dense Teacher with several existing methods under the COCO-Standard setting in Table.~\ref{table:2}. When the labeled data varies from 2\% to 10\%  our model consistently shows superior results. Whereas Under the 1\% labeled setting, the performance of Dense Teacher is lower than Faster R-CNN based Unbiased Teacher. However, a more direct comparison between our method and Unbiased Teacher under the 1\% setting on FCOS shows that Dense Teacher still leads by 1.2\% mAP. Moreover, when applying large-scale jittering for augmentation following the implementation of Soft Teacher, our method obtains more significant improvements and becomes new the state-of-the-art.
\subsubsection{VOC \& COCO-Full.}

\begin{table}[tb]
	\begin{center}
		\caption{Comparison with existing methods on Pascal VOC. Evaluations are performed on \texttt{VOC07 test}}
		\label{table:3}
		\scalebox{0.9}{
		$$
			\begin{tabular}{ccccc}
				\hline
				Method                            & Labeled                & Unlabeled                           & $AP_{50}$      & $AP_{50:95}$   \\
				\hline
				Supervised(Ours)                        & VOC07                  & None                                & 71.69            & 45.87            \\
				\hline
				CSD~\cite{csd}  & \multirow{7}{*}{VOC07} & \multirow{7}{*}{VOC12} & 74.7& - \\
				STAC~\cite{stac}                  &                        &                                     & 77.45          & 44.64          \\
				ISMT~\cite{meanteachers}&&& 77.23 & 46.23 \\
				Unbiased Teacher~\cite{ubteacher} &                        &                                     & 77.37          & 48.69          \\
				\emph{Li, et al.}~\cite{rethinkingp}  &   &  & 79.00 & 54.60\\
				Instant Teaching~\cite{instantteaching}                  &                        &                                     & 79.20          & 50.00          \\
				Ours                              &                        &                                     & \textbf{79.89} & \textbf{55.87} \\
				\hline
				CSD~\cite{csd} & \multirow{7}{*}{VOC07} & \multirow{7}{*}{\tabincell{c}{VOC12\\+\\COCO20cls}}  & 75.1 & - \\
				STAC~\cite{stac}                  &                        &                                     & 79.08          & 46.01          \\
				ISMT~\cite{meanteachers}&&& 77.75 & 49.59 \\
				Unbiased Teacher\cite{ubteacher}  &                        &                                     & 78.82          & 50.34          \\
				\emph{Li, et al.}~\cite{rethinkingp}  &   &  & 79.60 & 56.10\\
				Instant Teaching~\cite{instantteaching}                  &                        &                                     & 79.90          & 55.70          \\
				Ours                              &                        &                                     & \textbf{81.23} & \textbf{57.52} \\
				\hline
			\end{tabular}$$
		}
	\end{center}
\end{table}
Results in Table.~\ref{table:3} and Table.~\ref{table:4} show that Dense Teacher lead the performance in both settings. On VOC dataset, Dense Teacher improves its supervised baseline by 8.2\% and 10.0\% on AP50 and mAP (\emph{i.e.}, from AP50 to AP95). Under the COCO-Full setting, since the baseline reported in other works are not the same, we list the performance of each method in the form of ``baseline$\rightarrow$result''. Our approach obtains a boost of 3.1\% from the \texttt{2017unlabeled} set, which is much higher than CSD, STAC, and Unbiased Teacher. We finally apply the large-scale jittering trick and a longer training scheduler for a fair comparison with Soft Teacher, where Dense Teacher boosts mAP by 4.9\%, reaching 46.12\% mAP.
\setlength{\tabcolsep}{5pt}
\begin{table}[h]
	\begin{center}
		\caption{Experimental results on COCO-Full. Evaluations are done on COCO \texttt{val2017}. Note that 1x represents 90K training iterations, $N$x represents $N$x90K iterations. $\circledcirc$ means training with large scale jittering}
		\label{table:4}
		\scalebox{0.9}{
		\begin{tabular}{ccccc}
			\hline
			Method  & mAP                                       \\
			\hline
			CSD~\cite{csd} (3x)
			& $40.20\xrightarrow{-1.38}38.82$           \\
			STAC~\cite{stac}  (6x)
			& $ 39.48\xrightarrow{-0.27}39.21 $         \\
			ISMT~\cite{meanteachers}
			& $ 37.81\xrightarrow{+1.83}39.64 $         \\
			Instant-Teaching~\cite{instantteaching} 
			& $ 37.63\xrightarrow{+2.57}40.20 $         \\
			Unbiased Teacher~\cite{ubteacher} (3x)
			& $ 40.20\xrightarrow{+1.10}41.30 $         \\
			Humble Teacher~\cite{humbleteacher} (3x)
			& $ 37.63\xrightarrow{\textbf{+4.74}}42.37 $         \\
			\emph{Li, et al.}~\cite{rethinkingp} (3x)
			& $ 40.20\xrightarrow{+3.10}43.30 $         \\
			Ours(3x)   
			& $ 41.22\xrightarrow{+2.66}43.90$    \\
			Ours(6x)   
			& $ 41.22\xrightarrow{+3.70}\textbf{44.94}$ \\
			\hline
			$\rm{Soft~Teacher^\circledcirc}$~\cite{softteacher} (8x)
			& $ 40.90\xrightarrow{+3.70}44.60 $         \\
			$\rm{Ours(8x)^\circledcirc}$   
			& $ 41.24\xrightarrow{\textbf{+4.88}}\textbf{46.12} $             \\
			\hline
		\end{tabular}
		}
	\end{center}
\end{table}


\subsection{Ablation and Key Parameters}\label{sec4_4}


\subsubsection{Effect of Hard Negative Samples}
\label{sec:HN}
In Dense Teacher, hard negative samples/anchors can be better utilized. We explore three different strategies to study the impact of these hard negative regions\footnote{Since the ``unlabeled images'' under the COCO-Standard setting actually come with annotations, we can perform label assignments on images using these annotations. The difference between our division ($k=1$) and the assigned foreground is defined as hard negatives.}, including ``suppressing'', ``ignoring'' and ``selecting''. Results are shown in Table.~\ref{table:5}. We first suppress these samples to 0 and find a significant performance drop on both classification and regression branches compared to the original setting. Then, we ignore these samples when calculating loss. It turns out that this setting performs better than the ``suppress'' setting but still falls short when they are selected for training, indicating that learning to predict those hard negative samples can positively affect the model performance.

\subsubsection{Regression Branch} \label{sec:reg}
We have shown that unsupervised learning on the regression branch effectively improves model performance. However, since the output of deltas in the background region is not meaningful in FCOS, the quality of pseudo-labels in this branch is highly dependent on the design of the learning region. As can be seen in Table.~\ref{table:1}, in our region division and pseudo-box based method, the model can obtain gains of 1.8\% mAP and 1.6\% mAP. When using ground truth positives as learning region (see Table.~\ref{table:5} ``suppress'' and ``ignore''), the model gains about 2\% mAP. Therefore, our region division strategy can produce sufficiently reliable regions for this task.

\begin{table}[b]
\centering
\scalebox{0.95}{
    \begin{minipage}[!t]{0.4\columnwidth}
    		\caption{Impact of strategies dealing with hard negatives}
    		\label{table:5}
    		\centering
    		\begin{tabular}{cccccccccc}
    			\hline
    		   HN samples & Cls          & Reg          & AP   \\
    			\hline
                Suppress
    			       & $\checkmark$ & $\times$     & 31.56 \\
    		    Ignore
    			       & $\checkmark$ & $\times$     & 32.64 \\
    			Select
    			       & $\checkmark$ & $\times$     & \textbf{33.34} \\
    			\hline
                Suppress
    			       & $\checkmark$ & $\checkmark$ & 33.47 \\
    			Ignore
    			       & $\checkmark$ & $\checkmark$ & 34.72 \\
    			Select
    			       & $\checkmark$ & $\checkmark$ & \textbf{35.11} \\
    			\hline
    		\end{tabular}
    \end{minipage}
    \hspace{0.05\columnwidth}
    \begin{minipage}[!t]{0.5\columnwidth}
        \caption{Extensive comparison on different detectors and datasets}
    		\label{table:6}
    		\centering
    		\begin{tabular}{cccccccccc}
    			\hline
    		   Method & \small Anchor  & Dataset          & AP/MR   \\
    			\hline
                UT~\cite{ubteacher}&
    			      \checkmark & COCO & 28.9 \\
    		    Ours&
    		          \checkmark & COCO & \textbf{31.1}\\
    			\hline
    			UT~\cite{ubteacher}&
    			      $\times$ & CH\cite{crowdhuman} & 62.8 \\
    		    Ours &
    		          $\times$ & CH\cite{crowdhuman} & \textbf{60.7}\\
    			\hline
    		\end{tabular}
    \end{minipage}
}
\end{table}

\setlength{\tabcolsep}{4pt}
\begin{table}[tb]
	\caption{Ablation study on hyper-parameters introduced by our method. COCO 10\% stands for COCO-Standard 10\%}
	\label{table:7}
	\begin{center}
    \begin{tabular}{@{}cc@{}}
    \scalebox{0.9}{
		\subfloat[Size of learning region]{
        \begin{tabular}{ccccccc}
          \hline
        Setting & $k$(\%)  & AP & AP50 & AP75 
          \\
          \hline
          \multirow{5}{*}{\tabincell{c}{COCO\\10\%}}&0.1   & 25.76 &41.79 & 27.34 
          \\
          &0.5   & 34.11 &51.94 & 36.80 
          \\
         & 1     & \textbf{35.11} & 53.35 & 37.79
          \\
         & 3 & 34.47 &53.13&36.87
          \\
         & 5  & 33.85&53.00&36.03
          \\
          \hline
        \end{tabular}
    }
    } &
    \scalebox{0.9}{
		\subfloat[Unsupervised weight factor $w_u$]{
        \begin{tabular}{ccccccc}
          \hline
          Setting & $w_u$ & AP & AP50 & AP75 \\
          \hline
          \multirow{3}{*}{\tabincell{c}{COCO\\10\%}}&2  & 34.86 & 53.22 & 37.34  \\
          &4  & \textbf{35.11} & 53.35 & 37.79 \\
          &8  & 33.81 & 51.97 & 36.34 \\
          \hline
          \multirow{2}{*}{\tabincell{c}{COCO-\\Full}}&2  & \textbf{44.92} & 63.71 & 48.79 \\
          &4  & 43.04 & 61.60 & 46.87\\
          \hline
        \end{tabular}
		}
    }
    \end{tabular}
	\end{center}
\end{table}

\subsubsection{Effectiveness on other detectors and datasets}
Apart from comparison with state-of-the-arts, we also validate the effectiveness of our method on anchor-based detector and on CrowedHuman~\cite{crowdhuman} dataset. For anchor-based detector, we take RetinaNet as representation. When comparing with pseudo-box based Unbiased Teacher, our method stay ahead of the curve as show in~\ref{table:6}, Dense Teacher achieved a 2.2\% mAP improvement over Unbiased Teacher. On the CrowedHuman dataset with FCOS detector, our method obtains a 2.1\% mMR improvement as well.

\subsubsection{Size of Learning Region}
We compare Dense Teacher's performance under different selecting ratio $k$ in Table.~\ref{table:7}(a). As shown in the table, we obtain the best performance when selecting 1\% of the samples (5 layers of FPN) for unsupervised learning. According to statistics, there are about 0.4\% of positive samples in COCO under the label assigning rule of FCOS. Therefore, the optimal learning region not only contains ground truth positives, but also encompasses hard negatives that are valuable. This also suggests that although the model performance is affected by this hyperparameter, the statistical characteristics of the data set can help us determine the optimal value of this hyperparameter, mitigate model migration and deployment challenges.

\subsubsection{Unsupervised Loss Weight}
The weight of unsupervised data also has an important impact on the training results. The experimental results in~\ref{table:7}(b) turn out that: for a limited amount of supervised data like in the COCO-Standard 10\% setting, a relatively large weight of $4$ is favorable. Meanwhile, for the COCO-Full setting where much more labeled data are available, weight of $2$ is enough. We attribute this phenomenon to the different degrees of over-fitting. When only a small amount of annotations are available, a relatively large weight of unsupervised parts could introduce stronger supervision. In contrast, when given a large amount of labeled data, a small unsupervised weight could address and better utilize supervision.

\section{Conclusion}

In this paper, we revisit the form of pseudo-labels in existing semi-supervised learning. By analyzing various flaws caused by the lengthy pseudo-box generation pipeline, we point out that pseudo-box is a sub-optimal choice for unlabeled data. To address this issue, we propose the Dense Teacher, a SS-OD framework which adopts dense predictions from the teacher model as pseudo-labels for unlabeled data. Our approach is simpler but stronger. We demonstrate its efficacy by comparing Dense Teacher with other pseudo-box based SS-OD algorithms on MS-COCO and Pascal VOC benchmarks. Results on both benchmarks show that our Dense Teacher achieves state-of-the-art performance.

\clearpage
%
%
\bibliographystyle{splncs04}
\bibliography{denseteacher}
\end{document}